\documentclass[10pt,twocolumn,letterpaper]{article}

\usepackage{cvpr}              %
\usepackage[accsupp]{axessibility}

\usepackage{graphicx}
\usepackage{amsmath}
\usepackage{amssymb}
\usepackage{booktabs}
\usepackage{xspace}
\usepackage{multirow}
\usepackage[dvipsnames]{xcolor}
\usepackage{float}
\usepackage[numbers]{natbib}
\usepackage{makecell}

\usepackage[pagebackref,breaklinks,colorlinks]{hyperref}

\RequirePackage{enumitem}
\setlist[itemize]{nosep}

\renewcommand{\paragraph}[1]{\vspace{0.2em}\noindent \textbf{#1 \hspace{0.2em}}}

\usepackage[capitalize]{cleveref}
\crefname{section}{Sec.}{Secs.}
\Crefname{section}{Section}{Sections}
\Crefname{table}{Table}{Tables}

\usepackage[labelsep=period]{caption}
\RequirePackage{enumitem}
\setlist[itemize]{nosep}

\newcommand{\Fref}[1]{Figure~\ref{#1}}

\usepackage{xcolor}
\newcounter{todos}
\AtEndDocument{\ifnum\value{todos}>0 \PackageWarning{TODOS}{There are \arabic{todos} todos left in this paper! Fix them before submitting the paper!} \fi}

\newcommand{\Netname}{JIFF\xspace}
\newcommand{\NetnameFull}{Jointly-aligned Implicit Face Function for High Quality}

\newcommand{\thudataset}{THUman2.0\xspace}

\def\cvprPaperID{1979} %
\def\confName{CVPR}
\def\confYear{2022}

\begin{document}

\title{\Netname: \NetnameFull \\ Single View Clothed Human Reconstruction}

\author{Yukang Cao\textsuperscript{1}
\quad
Guanying Chen\textsuperscript{2}
\quad
Kai Han\textsuperscript{1}
\quad
Wenqi Yang\textsuperscript{1}
\quad
Kwan-Yee K. Wong\textsuperscript{1} 
\vspace{0.3em} 
\\
{\normalsize \textsuperscript{1}The University of Hong Kong} \qquad  
{\normalsize \textsuperscript{2}The Future Network of Intelligence Institute (FNii), CUHK-Shenzhen}
}

\maketitle

\begin{abstract}
This paper addresses the problem of single view 3D human reconstruction. Recent implicit function based methods have shown impressive results, but they fail to recover fine face details in their reconstructions. This largely degrades user experience in applications like 3D telepresence. In this paper, we focus on improving the quality of face in the reconstruction and propose a novel Jointly-aligned Implicit Face Function ({\bf \Netname}) that combines the merits of the implicit function based approach and model based approach. We employ a 3D morphable face model as our shape prior and compute space-aligned 3D features that capture detailed face geometry information. Such space-aligned 3D features are combined with pixel-aligned 2D features to jointly predict an implicit face function for high quality face reconstruction. We further extend our pipeline and introduce a coarse-to-fine architecture to predict high quality texture for our detailed face model. Extensive evaluations have been carried out on public datasets and our proposed JIFF has demonstrates superior performance (both quantitatively and qualitatively) over existing state-of-the-arts.
\end{abstract}

\section{Introduction}

Under the current social distancing measures of the COVID-19 pandemic, video conferencing has become the major form of daily communication. 
With the increased popularity of 3D hardware like AR goggles, 3D telepresence~\cite{orts2016holoportation} will soon likely emerge as the next generation communication standard. High quality 3D human reconstruction is at the core of this technology and is one of the current hottest topics. Traditional reconstruction methods depend on expensive capturing hardware and tedious calibration procedure to produce good looking models~\cite{Collet2015HighqualitySF}. This limits their applications to studio settings with expert users and greatly hinders the growth of AR/VR applications. It is highly desirable to develop easy-to-use tools that allow easy creation of high
\begin{figure}[h]
	\includegraphics[width=1.\linewidth]{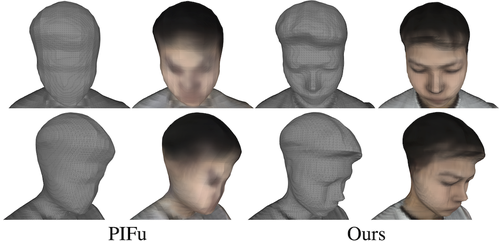}
	\caption{Reconstruction by PIFu~\cite{saito2019pifu} and our \Netname. Model reconstructed by \Netname shows much better geometric details and texture than that by PIFu.}
	\label{fig:first_page_result}
\end{figure}
quality 3D human models by home users using commodity RGB cameras.

With the advance in deep learning techniques, recent 3D human reconstruction methods have achieved impressive results using as few as a single image~\cite{saito2019pifu, jiang2020bcnet, zhu2019detailed}. These methods can be roughly divided into model based methods~\cite{bhatnagar2019mgn, alldieck2019tex2shape, 4270338, anguelov2005scape} and model-free methods~\cite{he2021arch++, chibane20ifnet, chen2018implicit_decoder, Peng2020ECCV, saito2019pifu, deng2020nasa}. 
Model based methods typically fit a parametric human model (\eg, SMPL~\cite{SMPL:2015}) to an image to produce a naked 3D human model. They have difficulties in recovering high-frequency details such as clothing and hair. Model-free methods, on the other hand, solve this problem by predicting the occupancy of a discretized volume space. One very representative model-free method is PIFu~\cite{saito2019pifu}, which exploits a Multi Layer Perceptron (MLP) to model an implicit function for predicting the occupancy value of a query point based on pixel-aligned features extracted from an image. PIFu and its variants \cite{saito2020pifuhd, li2020monocular} have achieved state-of-the-art results in free-form full body human reconstruction. However, their reconstructions are often lack of fine face details (see~\cref{fig:first_page_result}). Considering the ultra-high-definition rendering standards (\ie 4K UHD and 8K UHD) that are common nowadays, their face reconstruction quality is obviously far from satisfactory and largely degrades user experience in AR/VR applications like 3D telepresence.

To achieve high quality human reconstruction with fine face details, we propose a novel Jointly-aligned Implicit Face Function (aka \Netname) that combines the merits of the implicit function based approach and model based approach. Specifically, we employ the 3D morphable face model (3DMM)~\cite{blanz1999morphable} as our shape prior and compute space-aligned 3D features to capture detailed face geometry and texture information. There are also recent methods~\cite{he2020geopifu,zheng2021pamir} using 3D priors to enhance the implicit function representation by introducing geometric constraints to regularize the reconstruction. For example, \cite{he2020geopifu} and \cite{zheng2021pamir} utilize coarse 3D volume features and SMPL body model, respectively, to improve human body reconstruction. To the best of our knowledge, \Netname is the first method focusing on recovering high quality face details in both shape and texture.%

\Netname exploits space-aligned 3D features extracted from 3DMM as well as pixel-aligned 2D features extracted from image to jointly predict an implicit face function for high quality face reconstruction. In summary, our method first fits the 3DMM to the face in an image and employs two separate encoders to compute 3D shape and texture features, respectively, from the resulting 3D model. Given a 3D query point, we obtain its space-aligned 3D features with trilinear interpolation. Such space-aligned 3D features are combined with pixel-aligned 2D features for predicting the occupancy value of the query point using a MLP. We further extend our pipeline and introduce a coarse-to-fine architecture to predict high quality texture for our detailed face model.

By taking advantages of both the implicit function based approach and model based approach, our method can successfully recover fine face details in both shape and texture (see~\cref{fig:first_page_result}). Our key contributions are as follows:
\begin{itemize}[leftmargin=*]
    \item We propose \Netname, a novel implicit face function for high quality single view 3D face reconstruction, which integrates 3D face prior into the implicit function representation for high quality face shape reconstruction. 
    \item We exploit per-vertex color information provided by the 3DMM and introduce a coarse-to-fine architecture for high quality face texture prediction.
    \item We demonstrate how \Netname can naturally be extended to produce full body human reconstruction by simply appending a ``PIFu'' head (implemented as a MLP) to its convolutional image encoder.
    \item We carry out extensive experiments on public benchmarks and demonstrate that \Netname outperforms current state-of-the-arts by a large margin.
\end{itemize}

\section{Related work}
\paragraph{Single view human reconstruction}
Reconstructing 3D human body from a single image is an important and challenging problem which has attracted a considerable amount of attention.
Methods have been developed to fit parametric human models such as SCAPE~\cite{guan2009estimating}, SMPL~\cite{Bogo:ECCV:2016}, SMPL-X~\cite{pavlakos2019expressive} and STAR~\cite{osman2020star} to a single image. Human Mesh Recovery (HMR)~\cite{kanazawa2018end} proposes to regress SMPL parameters from a single image. Labels like body part segmentation~\cite{omran2018neural}, silhouette~\cite{pavlakos2018learning}, and IUV map~\cite{xu2019denserac} have been employed to provide intermediate supervisions for training SMPL prediction models. Although free-form deformation~\cite{joo2018total, xiang2019monocular, alldieck2018video, alldieck2019learning} may be applied to the models to partially account for complex shape topology (\eg, clothing and hair), these methods generally have difficulties in reconstructing high quality clothed human models.
Model-free methods have been proposed to reconstruct 3D human with arbitrary topology.
Voxel-based methods reconstruct 3D human body using a volumetric representation with different intermediate supervisions (\eg, multi-view images~\cite{gilbert2018volumetric}, 2D pose~\cite{varol2018bodynet, Zheng2019DeepHuman, Trumble_2018_ECCV}, and 3D pose~\cite{jackson20183d}). However, volumetric representation is memory intensive and is hard to scale to high resolution.
Recently, memory efficient implicit function representations~\cite{saito2019pifu, saito2020pifuhd, chen2018implicit_decoder, Wang2020IPNet, Peng2020ECCV, natsume2019siclope} have achieved outstanding performance in 3D reconstruction. 
DeepSDF~\cite{park2019deepsdf} predicts a signed distance field for surface reconstruction. IF-Net~\cite{chibane20ifnet} learns multi-scale features for 3D mesh refinement or completion, and is later extended for texture completion~\cite{chibane2020implicit}. SiCloPe~\cite{natsume2019siclope} reconstructs a visual hull by predicting 3D pose and 2D silhouettes. PIFu~\cite{saito2019pifu} introduces a pixel-aligned implicit function for human reconstruction from a single image. Variants have been proposed for high-resolution reconstruction~\cite{saito2020pifuhd} and real-time rendering~\cite{li2020monocular}. PeeledHuman~\cite{jinka2020peeledhuman} proposes to encode the human body as a set of peeled depth and RGB maps to handle the self-occlusion problem. Although these methods successfully reconstruct 3D human body geometry and texture from a single image, they fail to deal with issues like self-occlusion and detailed face reconstruction. 
\paragraph{3D prior for implicit function representation} 
Efforts have been made to utilize 3D human body prior for implicit model reconstruction.
GeoPIFu~\cite{he2020geopifu} proposes to use geometry-aligned feature from the predicted 3D feature volume to improve reconstruction quality. PaMIR~\cite{zheng2021pamir} utilizes SMPL as a 3D prior for implicit function learning. 
ARCH and ARCH++~\cite{Huang_2020_CVPR,he2021arch++} extract 3D spatially-aligned features from a canonical SMPL mesh to reconstruct animatable human models. SHARP~\cite{jinka2021sharp} proposes the peeled SMPL priors for learning. DeepMultiCap~\cite{zheng2021deepmulticap} uses SMPL to tackle the multi-person reconstruction problem. 
Although improved results have been reported, the above mentioned methods still cannot recover fine face details in their reconstructions. High quality face reconstruction remains an open problem.

\paragraph{Human face/head reconstruction} 
Parametric model based methods often adopt 3D Morphable Model (3DMM)~\cite{blanz1999morphable, guo2020towards, lin2020towards, accu-face-recons-19, egger20203d, sariyanidi2020inequality, zhu2017face} for face reconstruction 
and head models (\eg, FLAME~\cite{FLAME:SiggraphAsia2017}, DECA~\cite{DECA:Siggraph2021}, LYHM~\cite{8237597, Dai2019StatisticalMO}, and UHM~\cite{ploumpis2020towards, ploumpis2019combining}) for full-head reconstruction.
The models adopted by these methods, however, often limit their expressiveness in handling arbitrary shapes.
i3DMM~\cite{yenamandra2021i3dmm} combines 3DMM and implicit function to predict full-head meshes with fine details. 
H3D-Net~\cite{ramon2021h3d} optimizes the implicit function representation based on a 3D head model learned from thousands of raw scans.
Different from these methods, we aim at high quality single view full body human reconstruction with fine face details. 

\section{Preliminary}
In the following subsections, we are going to give a brief review on PIFu~\cite{saito2019pifu} and 3DMM~\cite{blanz1999morphable,accu-face-recons-19} which lay the foundation for \Netname.

\newcommand{\threeDx}{X}
\newcommand{\twoDx}{x}
\newcommand{\image}{I}
\newcommand{\implicitF}{f}
\newcommand{\pixelalignedF}{f_v}
\newcommand{\imagefeatureF}{\psi}
\newcommand{\twoDproject}{\pi}
\newcommand{\getdepth}{z}
\newcommand{\bilinear}{\mathcal{B}}
\newcommand{\trilinear}{\tau}
\newcommand{\facemeshS}{\mathbf{S}}

\subsection{Pixel-aligned implicit function}
\label{sec:pifu}
\begin{figure}[t]
  \centering
  \includegraphics[width=1\linewidth]{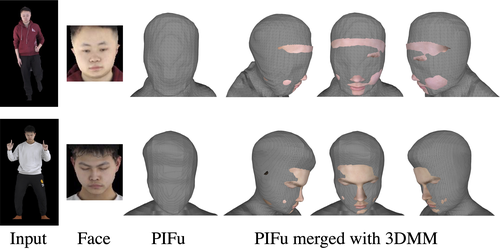}
   \caption{PIFu cannot recover fine face details in its reconstructions. 3DMM with detailed geometry can be fitted to the image, but it is a non-trivial task to merge the 3DMM mesh with the mesh extracted from PIFu.}
   \label{fig:preliminary}
\end{figure}

Recently, implicit functions~\cite{sclaroff1991generalized} have been widely adopted for 3D reconstruction~\cite{mescheder2019occupancy}. Denoting $\threeDx \in \mathbb{R}^3$ as a 3D point, a deep implicit function, modeled by a MLP, defines a surface as the level set of the function, \eg, $\implicitF(X)=0.5$.
The pixel-aligned implicit function $\pixelalignedF$ introduced in~\cite{saito2019pifu} is written as
\begin{equation}\label{PIFufeature}
    \pixelalignedF(\bilinear(\imagefeatureF(\image), \twoDproject(\threeDx)), z(\threeDx)) \mapsto [0, 1] \in \mathbb{R},
\end{equation}
where $\getdepth(\threeDx)$ is the depth of $\threeDx$ and $\bilinear(\imagefeatureF(\image), \twoDproject(\threeDx))$ is the pixel-aligned feature, with $\imagefeatureF(\image)$ denotes the feature map extracted from the image $\image$ by a convolutional encoder~\cite{newell2016stacked}, $\twoDproject(\threeDx)$ the 2D projection of $\threeDx$ on $\image$, 
and $\bilinear(\cdot)$ a bilinear interpolation operation. Despite its simplicity, PIFu has achieved impressive results in full body human reconstruction. However, PIFu is incapable of recovering fine face details in its reconstructions (see~\cref{fig:preliminary}).

\subsection{3DMM as 3D face prior}
\label{sec:param}
A natural idea to improve PIFu is to enforce some sort of 3D shape prior in learning the implicit function representation.
Efforts have been made to enhance the pixel-aligned implicit function with 3D features like coarse 3D volume features~\cite{he2020geopifu} and voxelized SMPL mesh features~\cite{zheng2021pamir, Huang_2020_CVPR, he2021arch++}. While improved overall reconstruction results have been reported, these method still cannot recover fine face details in their reconstructions. This is not unexpected as the shape priors adopted by these methods do not actually focus on the face.

An ideal 3D face prior should provide both geometry and texture information, and can be robustly estimated from an image. Parametric face/head models like 3DMM~\cite{blanz1999morphable} and DECA~\cite{DECA:Siggraph2021} are potential good candidates as they provide both 3D mesh and texture information to resolve depth ambiguity and improve texture prediction. They can be delineated with a small number of parameters, which can be estimated effectively using existing methods (\eg~\cite{accu-face-recons-19}). In this work, we choose the widely used 3DMM as our 3D face prior for its simplicity and efficacy. 
Note that other parametric face models can also be adopted as \Netname does not depend on a specific parametric model.

\begin{figure*}[t]
  \centering
   \includegraphics[width=1\linewidth]{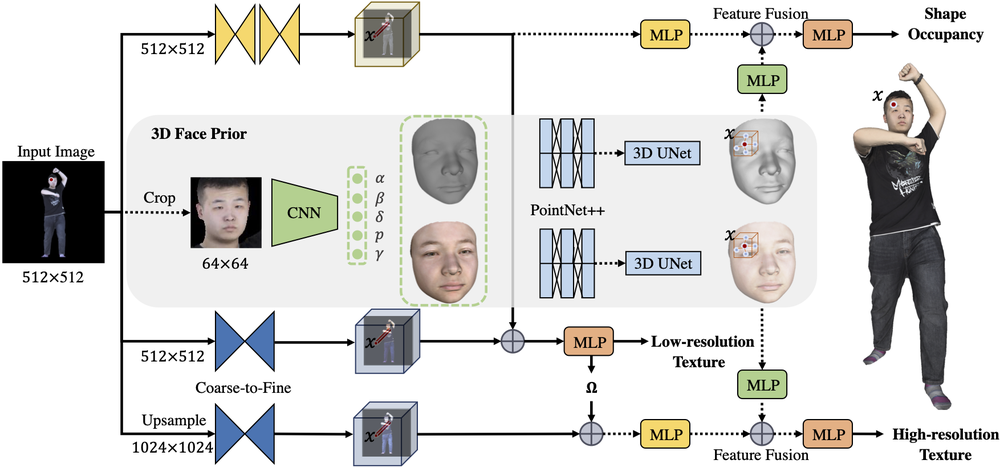}

   \caption{Overview of the network architecture for our proposed Jointly-aligned Implicit Face Function (\Netname). 
   JIFF takes a single image as input to reconstruct the high quality 3D human model with fine face details. It seamlessly incorporates 3D face prior into the implicit function representation for both shape and texture reconstruction. By appending two MLP heads, one after the pixel-aligned feature (in parallel with the MLP in yellow) in the top branch and the other after the multi-scale pixel-aligned feature (in parallel with the MLP in yellow) in the bottom branch, JIFF can be naturally extended to produce full body reconstruction. The dashed paths are unique for the face region. The rest are shared for the full body reconstruction. 
}
   \label{fig:pipeline}
\end{figure*}

3DMM models a face as a linear combination of the Principle Component Analysis (PCA) basis vectors.
The shape $\mathbf{S}$ and texture $\mathbf{T}$ of 3DMM are expressed as 
\begin{equation}\label{eq:3dmm-para}
\begin{split}
    \mathbf{S}&=\bar{\mathbf{S}}+\mathbf{B}_{id} \boldsymbol{\alpha}+\mathbf{B}_{exp} \boldsymbol{\beta}, \\
    \mathbf{T}&=\bar{\mathbf{T}}+\mathbf{B}_{tex} \boldsymbol{\delta},
\end{split}
\end{equation}
where $\bar{\mathbf{S}}$ and $\bar{\mathbf{T}}$ denote the mean shape and texture respectively, $\mathbf{B}_{id}$, $\mathbf{B}_{exp}$, and $\mathbf{B}_{tex}$ are the PCA bases for face identity, expression, and texture respectively, 
and $\boldsymbol{\alpha}$, $\boldsymbol{\beta}$, and $\boldsymbol{\delta}$ are the identity, expression, and texture coefficients respectively. In our implementation, we employ $\bar{\mathbf{S}}$, $\bar{\mathbf{T}}$, $\mathbf{B}_{id}$, and $\mathbf{B}_{tex}$ extracted from BFM~\cite{paysan20093d}, and $\mathbf{B}_{exp}$ built from FaceWarehouse~\cite{cao2013facewarehouse}. 

\newcommand{\fepifuF}{f}
\newcommand{\facepriorF}{\varphi}

\section{Jointly-aligned implicit face function}
\label{sec:faceenhance}
In this section, we introduce our novel Jointly-aligned Implicit Face Function (\Netname). \Netname is designed with the goal of incorporating 3D face prior in learning implicit function for high quality single view clothed human reconstruction. As already mentioned in the previous section, we choose 3DMM as our 3D face prior for its simplicity and efficacy. Given an input image, we can obtain the 3DMM parameters using~\cite{accu-face-recons-19}. However, merging the resulting 3DMM mesh with the mesh extracted from PIFu is a non-trivial problem, as they do not share the same geometry and topology (see~\cref{fig:preliminary}). Instead of trying to merge the two meshes na\"ively in the 3D space, we propose to fuse the information in the feature space, and use both space-aligned 3D features extracted from 3DMM and pixel-aligned 2D features extracted from image to jointly estimate an implicit face function (and hence the name Jointly-aligned Implicit Face Function).

Given a 3DMM mesh $\facemeshS$ fitted to the input image, we employ an encoder to generate a feature volume $\facepriorF(\facemeshS)$ from the mesh (see~\cref{subsec:volumeencoder} for details). \Netname can be formulated as
\begin{equation}\label{eq:FEPIFu}
    \fepifuF^g(
    \trilinear(\facepriorF(\facemeshS), \threeDx), 
    \bilinear(\imagefeatureF(\image), \twoDproject(\threeDx)),
    z(\threeDx)) \mapsto [0, 1] \in \mathbb{R},
\end{equation}
where $\trilinear(\facepriorF(\facemeshS), \threeDx)$ is the space-aligned 3D feature for the query point $\threeDx$ obtained by trilinear interpolation $\trilinear$. By employing both pixel-aligned 2D features and space-aligned 3D features, \Netname enjoys the benefits of both the implicit function based approach and model based approach and is capable of recovering fine face details in its reconstructions.

In the following subsections, we are going to give the details for each of the components in our \Netname learning framework. We are also going to describe how \Netname can naturally be extended for full body reconstruction by simply appending a ``PIFu'' head (implemented as a MLP) to its convolutional image encoder.

\subsection{3DMM prediction and alignment}
Given an input image, we first detect the face region~\cite{zhang2016joint} and adopt the state-of-the-art 3DMM regression method~\cite{accu-face-recons-19} to predict the 3DMM parameters $(\boldsymbol{\alpha}, \boldsymbol{\beta}, \boldsymbol{\delta}, \boldsymbol{\gamma}, \boldsymbol{p}) \in \mathbb{R}^{257}$, where $\boldsymbol{\alpha} \in \mathbb{R}^{80}$, $\boldsymbol{\beta} \in \mathbb{R}^{64}$, $\boldsymbol{\delta} \in \mathbb{R}^{80}$, $\boldsymbol{\gamma} \in \mathbb{R}^{27}$, and $\boldsymbol{p} \in \mathbb{R}^{6}$ represent face identity, expression, texture, illumination, and pose respectively. We can then obtain the 3DMM mesh and texture using \cref{eq:3dmm-para}, and apply the pose parameters to transform the mesh into the camera coordinate system. Next, we scale the 3DMM mesh to match the original size of the face in the input image\footnote{Scaled and cropped face images are used in the 3DMM regression.}, and further align the mesh with the back-projected face landmarks. At training time, we carry out the Iterative Closest Point (ICP)~\cite{arun1987least} algorithm to align the 3DMM mesh with the ground-truth mesh instead. This step is important in computing reliable space-aligned 3D features for training and testing. Details for the alignment processes (both for training and testing) can be found in the supplementary material.

\newcommand{\threeDgeometryF}{\facepriorF_g}
\newcommand{\threeDtextureF}{\facepriorF_t}

\subsection{Point-based 3D face feature encoding}
\label{subsec:volumeencoder}
To extract expressive features from the 3DMM mesh, we need to have a proper feature encoder.
Inspired by~\cite{Peng2020ECCV}, which predicts occupancy from point clouds, we adopt PointNet++~\cite{qi2017pointnet++} to extract hierarchical 3D point features for each vertex in the 3DMM mesh. These 3D point features are then projected into a 3D feature volume using average pooling. This feature volume is further processed by 3D U-Net~\cite{cciccek20163d} to aggregate local and global information.
In our implementation, the spatial dimension of our 3D feature volume is set to $64 \times 64 \times 64$, and the features extracted by 3D U-Net have a dimension of 128. Instead of using a single feature encoder consisting of PointNet++ and 3D U-Net, we propose to employ two separate encoders in our framework: one focusing on shape prediction by taking the plain 3DMM mesh as input, and the other focusing on texture prediction by taking the textured 3DMM mesh as input (see~\cref{fig:pipeline}).
The 3D features extracted by the geometry and texture encoders are denoted as $\threeDgeometryF(\facemeshS)$ and $\threeDtextureF(\facemeshS)$ respectively. Without any color information, $\threeDgeometryF(\facemeshS)$ is forced to learn features that depend only on geometry, whereas the additional color information helps learn better texture in $\threeDtextureF(\facemeshS)$.

\subsection{Occupancy prediction}
Referring to \cref{eq:FEPIFu}, \Netname takes both space-aligned 3D geometry features $\trilinear(\facepriorF_g(\facemeshS), \threeDx)$ and pixel-aligned 2D features $\bilinear(\imagefeatureF_g(\image), \twoDproject(\threeDx))$ to predict the occupancy value of a query point \threeDx. Following~\cite{saito2019pifu}, we adopt stack-hourglass encoder~\cite{newell2016stacked} for image feature extraction. The image encoder $\imagefeatureF_g$ takes image $I \in \mathbb{R}^{512 \times 512 \times 3}$ as input, with background masked out by the segmentation mask and produces a feature map $\imagefeatureF_g(I) \in \mathbb{R}^{128 \times 128 \times 256}$. The pixel-aligned 2D feature for $\threeDx$ is then obtained via bilinear interpolation on $\imagefeatureF_g(I)$. Similarly, the space-aligned 3D geometry feature for $\threeDx$ is obtained via trilinear interpolation on $\facepriorF_g(\facemeshS)$.

Inspired by Geo-PIFu~\cite{he2020geopifu}, instead of directly concatenating the pixel-aligned 2D feature and space-aligned 3D geometry feature, we first apply two separate MLPs to transform these features independently. The transformed features are then concatenated and fed into an MLP for occupancy prediction (see~\cref{fig:pipeline}). We train our shape prediction network by minimizing the mean squared error between the predicted and ground-truth occupancy values of the query points. The ground-truth occupancy value is 1 if $\threeDx$ is inside the surface, and 0 otherwise.

\begin{table*}[t]
\centering
\caption{Quantitative comparison on head/face and body-only reconstructions. Results are measured in $cm$ (the lower the better).
}
\label{tab:quantitative}
\resizebox{1.\textwidth}{!}{

\begin{tabular}{lcccccccccc}

\toprule
{} & \multicolumn{6}{c}{Head/Face region} & \multicolumn{4}{c}{Body-only region} \\
\cmidrule[0.5pt](rl){2-7}
\cmidrule[0.5pt](rl){8-11}
{} & \multicolumn{3}{c}{THuman2.0}  & \multicolumn{3}{c}{BUFF} & \multicolumn{2}{c}{THuman2.0}  & \multicolumn{2}{c}{BUFF} \\
\cmidrule[0.5pt](rl){2-4}
\cmidrule[0.5pt](rl){5-7}
\cmidrule[0.5pt](rl){8-11}
{Method} & {Face $L2$ distance$\downarrow$} & {Head P2S$\downarrow$} & {Head Chamfer$\downarrow$} & {Face $L2$ distance$\downarrow$} & {Head P2S$\downarrow$} & {Head Chamfer$\downarrow$}& {P2S$\downarrow$} & {Chamfer$\downarrow$} & {P2S$\downarrow$} & {Chamfer$\downarrow$}\\
\midrule
PIFu~\cite{saito2019pifu}  & 0.427 & 0.761 & 0.756 & 0.462 & 0.863 & 0.897 & 1.747 & 1.768 & 1.883 & 1.971 \\         
PIFuHD~\cite{saito2020pifuhd}  & 0.650 & 0.855 & 0.907 & 0.711 & 0.975 & 1.048 & \textbf{1.459} & \textbf{1.526} & \textbf{1.690} & \textbf{1.774} \\
PaMIR~\cite{zheng2021pamir}  & 0.403 & 0.693 & 0.714 & 0.447 & 0.805 & 0.819 & 1.607 & 1.617 & 1.751 & 1.804 \\
Ours  & \textbf{0.141} & \textbf{0.291} & \textbf{0.308} & \textbf{0.190} & \textbf{0.389} & \textbf{0.412} & 1.685 & 1.706 & 1.811 & 1.893 \\
\bottomrule
\end{tabular}
}
\end{table*}

\subsection{Face texture prediction}
Apart from recovering geometric details for face reconstruction, \Netname is also capable of improving face texture prediction.
Similar to shape prediction, the deep implicit function for face texture prediction can be formulated as
\begin{equation}\label{eq:FEPIFuTexture}
    \fepifuF^t(
    \trilinear(\facepriorF_t(\facemeshS), \threeDx), 
    \bilinear(\imagefeatureF_t(\image), \twoDproject(\threeDx)),
    z(\threeDx)) \in \mathbb{R}^3,
\end{equation}
where $\trilinear(\facepriorF_t(\facemeshS), \threeDx)$ is the space-aligned 3D texture feature and $\bilinear(\imagefeatureF_t(\image), \twoDproject(\threeDx))$ is the pixel-aligned 2D feature. The space-aligned 3D texture feature extracted by the texture feature encoder $\facepriorF_t(\facemeshS)$ embeds space-aware texture information from the textured 3DMM, which is particularly helpful for face texture prediction. Note that none of the existing methods (\eg PAMIR~\cite{zheng2021pamir}) that employ parametric models to improve human reconstruction consider such space-aware texture information due to the absence of texture information in their parametric models. 

\paragraph{Coarse-to-fine texture prediction}
Thanks to the space-aligned 3D features extracted from the textured 3DMM mesh, \Netname can produce better face texture than PIFu. To further improve the quality of the predicted texture, we, inspired by H3D-Net~\cite{ramon2021h3d}, introduce a coarse-to-fine architecture for texture prediction that exploits pixel-aligned features extracted from a higher resolution image (see~\cref{fig:pipeline}). Our proposed architecture is composed of a coarse branch and a fine branch. The coarse branch is identical to Tex-PIFu~\cite{saito2019pifu} which takes the concatenation of the pixel-aligned feature from the shape prediction network and pixel-aligned feature from the texture prediction network as input to a MLP to predict coarse texture. 
The fine branch takes an upsampled image of size $1024\times 1024\times 3$ as input to the image encoder to produce a feature map of size $512\times512\times256$ (4 times the width and height of the feature map in the coarse branch). Pixel-aligned feature computed from this fine feature map is concatenated with the output from the penultimate layer (denoted as $\Omega$ in~\cref{fig:pipeline}) of the MLP in the coarse branch to form a {\em multi-scale pixel-aligned feature}. Similar to the shape prediction network, this multi-scale pixel-aligned feature and the space-aligned texture feature are transformed independently by two separate MLPs before they are being concatenated and fed to the final MLP for fine texture prediction.

We first train the coarse branch, and then train the fine branch with the parameters of coarse branch being frozen. $L1$ loss is used to train both the coarse and fine branches. At training time, we randomly perturb $\threeDx$ with an offset $\epsilon \in \mathcal{N}(0, d)$ along its unit surface normal $N$, \ie $\threeDx^{'} = \threeDx + \epsilon \cdot N$, for point sampling~\cite{saito2019pifu}. This strategy allows the color of a surface point to be defined in a 3D space around its exact location which can stabilize the training process.

\subsection{Full body reconstruction}
Up till now, our discussion has been focused on face reconstruction. \Netname can actually be naturally extended for full body reconstruction by simply appending a ``PIFu'' head (implemented as a MLP) to its convolutional image encoder in the shape prediction network. Given a query point $\threeDx$, we apply \Netname to predict its occupancy value if it is projected to the face region in the input image, otherwise we predict its occupancy value using the ``PIFu'' head. Adopting other PIFu variants for improved reconstruction is also trivial. Similarly, to predict the texture for a non-face query point, we append a ``non-face texture'' head (also implemented as a MLP) to the texture prediction network which takes the \emph{multi-scale pixel-aligned feature} as input to predict the texture color. Though without the space-aligned 3D features, our coarse-to-fine design can also notably improve the texture prediction for non-face regions.

\section{Experiments}
In this section, we evaluate our proposed \Netname on public datasets and compare it with other state-of-the-art methods. 

\paragraph{Implementation details} 
We apply the stack-hourglass image encoder \cite{newell2016stacked} to extract image feature for shape prediction. Following the design of Peng~\etal \cite{Peng2020ECCV}, our 3D point feature encoders are composed of a PointNet++ and a 3D-UNet, and they produce 3D feature volume of size $64 \times 64 \times 64 \times 128$. We employ MTCNN \cite{zhang2016joint} to detect and crop the face region from the input image, and adopt the model proposed by Deng~\etal \cite{accu-face-recons-19} with a ResNet50 backbone for 3DMM parameter prediction. We implement the ResNet module from CycleGAN generator~\cite{zhu2017unpaired} as the image feature encoder for texture prediction.
We first train the shape prediction network for $9$ epochs with a learning rate of $0.0001$, which is frozen afterwards. We then train the coarse and fine branches of the texture prediction network sequentially for $6$ epochs each with a learning rate of $0.001$. Note that we freeze the coarse branch while training the fine branch. The batch size is set to $3$ to train both branches. We train our networks with the RMSprop \cite{ruder2016overview} optimizer. 
During training, we sample $5,700$ points in the 3D space as the query points for each input image, where $5,000$ points are sampled in the full body region and $700$ points are sampled uniformly in the face region. To sample the $5,000$ query points in the body region, we follow Saito~\etal \cite{saito2019pifu} to uniformly sample $15/16$ of the points on the mesh surface followed by a Gaussian perturbation along the surface normal direction, and uniformly sample 
the remaining $1/16$ of the points within the bounding box of the mesh. This strategy is helpful to eliminate isolated outliers in the reconstruction. We implement our method using PyTorch~\cite{paszke2017pytorch} and carry out our experiments on three NVIDIA 2080Ti GPUs.
Our code will be made publicly available at \url{https://yukangcao.github.io/JIFF}.

\paragraph{Datasets}
Many recent methods use RenderPeople\footnote{\url{https://renderpeople.com}} and AXYZ datasets\footnote{\url{https://secure.axyz-design.com}} to train their models. However, these two datasets are commercial data and not publicly available. Instead, we use the public \thudataset dataset~\cite{tao2021function4d} as our main testbed, which contains high-quality human scans with different clothes and poses, captured by a dense DSLR rig. 
3D mesh and the corresponding texture map are provided for each subject. 
We randomly split the data into 465 subjects for training and 61 subjects for testing.
For each subject, we follow Saito~\etal \cite{saito2019pifu} to render $360^\circ$ images by rotating the camera around the mesh and varying the illumination. As our main focus is to improve face details in the reconstruction, we omit images in which no human face can be detected. 
In addition, we use BUFF dataset~\cite{Zhang_2017_CVPR}, which contains 5 subjects, as an additional testing dataset to evaluate our method. Besides, we evaluate \Netname on 2 free models from RenderPeople.

\begin{figure*}[h]
	\includegraphics[width=1\linewidth]{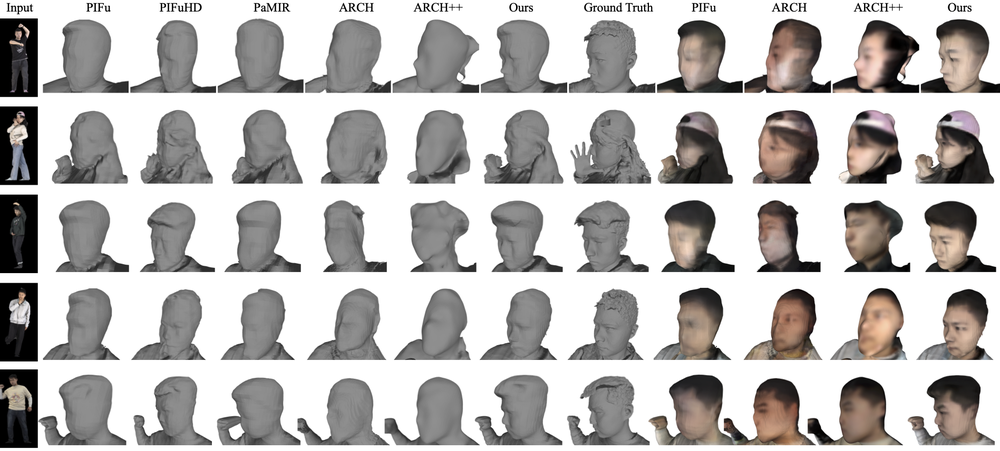}
	\caption{Qualitative comparison with other state-of-the-arts. Here we focus on comparing the face geometry and texture. Please refer to the supplementary material for full body reconstruction results.}
	\label{fig:overall comparison}
\end{figure*}

\subsection{Comparison with state-of-the-arts}
We compare our method with recent state-of-the-art methods, including PIFu~\cite{saito2019pifu}, PIFuHD~\cite{saito2020pifuhd}, PaMIR~\cite{zheng2021pamir}, ARCH~\cite{Huang_2020_CVPR}, and ARCH++~\cite{he2021arch++}. 
For a fair comparison, we retrain PIFu, PIFuHD, and PaMIR on the \thudataset, as they are originally trained on either non-public commercial data or a different version of THUman data.
As the codes for ARCH and ARCH++ are not publicly available, their authors help to provide the evaluation results on our data upon our request.

\begin{figure}[h]
	\includegraphics[width=1\linewidth]{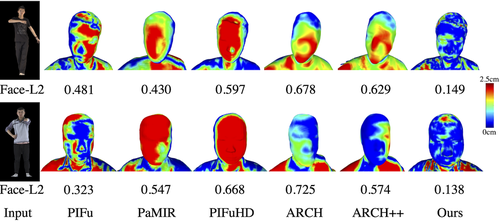}
	\caption{Error map of the face region based on P2S distance. Our proposed method achieves much better results compared with other state-of-the-arts.}
	\label{fig:error_color}
\end{figure}

\paragraph{Quantitative comparison}
We quantitatively evaluate on the test split of \thudataset dataset and the BUFF dataset~\cite{Zhang_2017_CVPR}. 
As our main focus is to recover faces with fine details, we first compare our method with others on the head region. The results are shown in~\cref{tab:quantitative}.
We measure both Point-to-Surface distance (P2S) and Chamfer mean distance in the head region and $L2$ face mean distance~\cite{ramon2021h3d} in the face region by back-projecting 2D image points to 3D face surface. It can be observed that our method significantly outperforms all the others on both datasets, confirming that our method is capable of recovering fine face details. We locate the head region by applying an off-the-shelf human parsing method~\cite{zhao2020human}. It is interesting to note that PIFuHD performs the worst for the head/face region among the methods under comparison.
We further show body-only reconstruction results in ~\cref{tab:quantitative} and report the P2S and Chamfer distances. We can observe that our \Netname slightly improves the body-part quality from PIFu, although there is no 3DMM-like extra information adopted for training the body. We think the improvement originates from the fact that JIFF is jointly trained on face and body. The shared network backbone is updated by gradients from both head and body branches and the improved head branch also affect the body branch in a positive way, resulting in a stronger backbone for both head and body reconstruction.

\begin{figure}[h]
	\includegraphics[width=1\linewidth]{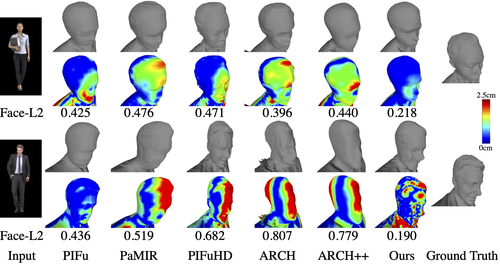}
	\caption{Reconstruction results for two RenderPeople subjects. JIFF is able to reconstruct the face details that others struggle.}
	\label{fig:rp-testing}
\end{figure}

\paragraph{Qualitative comparison}
We qualitatively compare our method with the state-of-the-arts in~\cref{fig:overall comparison} for both face shape and texture (the corresponding full body reconstructions can be found in the supplementary material). Our method can faithfully recover fine face details in terms of both shape and texture.
We successfully reconstruct the geometry of the nose, mouth, and eyes. The overall shape of our reconstruction is also closest to the ground truth. Other methods struggle in these aspects. We show the error map in~\cref{fig:error_color} which demonstrates that our method outperforms others for the majority part of the face region. We also evaluate our method on two free models from RenderPeople in~\cref{fig:rp-testing}. Again, our method is capable of recovering fine face details, which is consistent with the ground truth. 

\subsection{Ablation study}
\label{sec:ablation}
\begin{table}
  \centering
\caption{Effectiveness of our 2D-3D feature fusion. ($\oplus$ denotes the concatenation operator.) 
}
\label{tab:quantitative_mlp}
\resizebox{0.48\textwidth}{!}{
  \begin{tabular}{lccc}
  \toprule
  {} & Face $L2$ distance $\downarrow$ & Head P2S $\downarrow$ & Head Chamfer $\downarrow$ \\
\midrule
(a) 2D only & 0.427 & 0.761 & 0.756 \\
(b) 3D only & 0.279 & 0.485 & 0.492 \\
(c) 2D $\oplus$ 3D & 0.171 & 0.310 & 0.332 \\
(d) MLP(2D)  $\oplus$ MLP(3D) & \textbf{0.141} & \textbf{0.291} & \textbf{0.308} \\ 
\bottomrule
\end{tabular}
}
\end{table}
\paragraph{2D-3D feature fusion} 
We validate the effectiveness of the fusion of pixel-aligned 2D feature with space-aligned 3D feature in our framework by comparing the following variants: (a) using pixel-aligned 2D feature alone; (b) using  space-aligned 3D feature alone; (c) simply concatenating the 2D and 3D features; and (d) applying transformation by additional MLPs before concatenation (\ie implementation of \Netname). The intuition of applying the additional MLPs before concatenation is to learn a proper embedding space for the two different types of features to fuse better. The results are reported in~\cref{tab:quantitative_mlp}. It can be observed that \Netname significantly outperforms the other variants.

\begin{figure}
\includegraphics[width=1\linewidth]{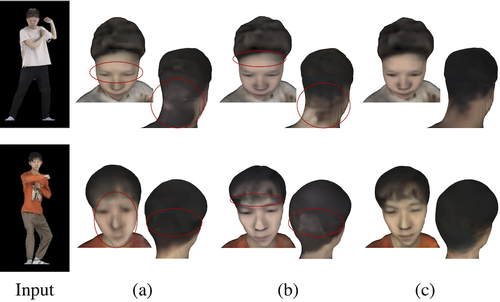}
\caption{Reconstruction results for different 3D feature encoder designs. From left to right: (a) Single encoder with plain 3DMM. (b) Single encoder with textured 3DMM. (c) Dual encoders with plain 3DMM and textured 3DMM.}
\label{fig:ablation_encoder}
\end{figure}

\begin{figure}
\includegraphics[width=1\linewidth]{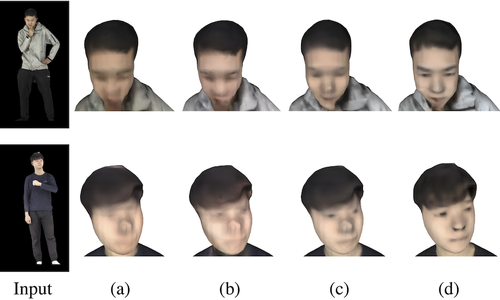}
\caption{Ablation study on different architectures for texture prediction. From left to right: (a) coarse resolution only (original PIFu); (b) fine resolution only; (c) fine resolution w/ 3D face prior; (d) coarse-to-fine w/ 3D face prior (ours).}
\label{fig:ablation_texture}
\vspace{-0.2cm}
\end{figure}

\paragraph{Dual 3D feature encoders}
In our framework, we introduce two separate 3D feature encoders focusing on improving the shape and texture details respectively. They share the same architecture but take different 3DMM meshes as input, namely the plain 3DMM mesh and the textured 3DMM mesh. To validate the effectiveness of our design, we compare the following variants:  (a) a single 3D feature encoder taking the plain 3DMM mesh as input; (b) a single 3D feature encoder taking the textured 3DMM mesh as input; and (c) dual 3D feature encoders taking the plain and textured 3DMM meshes as input respectively (\ie the implementation of \Netname). We report SSIM \cite{wang2004image}, LPIPS \cite{zhang2018perceptual}, and $L1$ error in \cref{tab:texture_encoder}. SSIM and LPIPS are measured by reprojecting the reconstructed 3D model onto the image and compare the performance of the face region. We employ the same method as in \cite{zhao2020human} again to parsing the 3D reconstruction and obtain the head region followed by calculating the normalized $L1$ vertex color error. ~\Fref{fig:ablation_encoder} shows some qualitative results for the three variants. It can be observed that our dual encoder design achieves the best performance for both shape and texture reconstructions. The first two variants would result in blurry face texture or poor backside texture. Moreover, our dual encoder successfully help improve the texture of the backside of the head.

\paragraph{Coarse-to-fine texture prediction}
In~\cref{tab:texture_training} and~\cref{fig:ablation_texture}, we compare our coarse-to-fine architecture for texture prediction in our framework with the following variants: (a) coarse resolution only; (b) fine resolution only; (c) fine resolution jointly with 3D face prior; and (d) coarse-to-fine setting jointly with 3D face prior. From both qualitative and quantitative results, fine resolution and 3D face prior could indeed help gain high-resolution texture, and face texture details respectively. Also, it can be observed that our coarse-to-fine architecture can notably improve the texture prediction over its single resolution counterparts.

\begin{table}
  \centering
\caption{Effectiveness of dual 3D feature encoders.}
\label{tab:texture_encoder}
\resizebox{0.48\textwidth}{!}{
  \begin{tabular}{lccc}
  \toprule
  {} & \makecell{Face SSIM $\uparrow$} & Face LPIPS $\downarrow$  & Head $L1$ error $\downarrow$   \\
\midrule
(a) single encoder + plain 3DMM & 0.7563 & 0.1143 & 0.1137 \\
(b) single encoder + textured 3DMM & 0.7589 & 0.1122 & 0.1090 \\
(c) dual encoders + plain/textured 3DMM & \textbf{0.7649} & \textbf{0.1107} & \textbf{0.1051} \\
\bottomrule
\end{tabular}
}
\end{table}

\begin{table}
  \centering
\caption{Effectiveness of our coarse-to-fine architecture for texture prediction.}
\label{tab:texture_training}
\resizebox{0.48\textwidth}{!}{
  \begin{tabular}{lccc}
  \toprule
  {} & \makecell{Face SSIM $\uparrow$} & Face LPIPS $\downarrow$  & Head $L1$ error $\downarrow$   \\
\midrule
(a) Coarse & 0.7126 & 0.1281 & 0.1346 \\
(b) Fine & 0.7264 & 0.1254 & 0.1285 \\
(c) Fine w/ 3D & 0.7580 & 0.1126 & 0.1097 \\
(d) Coarse-to-Fine w/ 3D & \textbf{0.7649} & \textbf{0.1107} & \textbf{0.1051} \\
\bottomrule
\end{tabular}
}
\end{table}

\section{Conclusion}
We have presented \Netname, a novel Jointly-aligned Implicit Face Function for high quality single view 3D human reconstruction, by incorporating the 3D face prior, in the form of 3DMM, into the implicit representation.  We further introduced a coarse-to-fine architecture for high quality face texture reconstruction. By simply appending two MLPs, one for shape and the other for texture, to JIFF, we successfully extended it to produce full body human reconstruction. We thoroughly evaluated JIFF on public benchmarks, establishing the new state-of-the-art for face details reconstruction. As JIFF is simple and effective, we believe it can be easily coupled with other human reconstruction approaches to improve their reconstruction quality for faces. 
Though \Netname achieves better performance than existing methods, subtle face details like eyelids still cannot be accurately reconstructed. One future research direction therefore is improving the quality for subtle details, as well as extending the idea to other body parts, \eg, hands and feet.

\paragraph{Acknowledgements}
This work was partially supported by Hong Kong RGC GRF grant (project\# 17203119), the National Key R\&D Program of China (No.2018YFB1800800), and the Basic Research Project No.~HZQB-KCZYZ-2021067 of Hetao Shenzhen-HK S\&T Cooperation Zone. We thank Yuanlu Xu for ARCH and ARCH++ results. 

{\small
\bibliographystyle{ieee_fullname}
\bibliography{PaperForReview}
}

\end{document}